\documentclass[fleqn,10pt]{wlscirep}
\usepackage[utf8]{inputenc}
\usepackage[T1]{fontenc}
\title{Data-driven Methods of Extracting Text Structure and Information Transfer}

\author[1]{Shinichi Honna}
\author[2]{Taichi Murayama}
\author[3,*]{Akira Matsui}

\affil[1]{Graduate School of Advanced Science and Technology, Japan Advanced Institute of Science and Technology}
\affil[2]{Graduate School of Environmental Information Studies, Yokohama National University}
\affil[3]{Center for Computational Social Science, Kobe University}
\affil[*]{amatsui@rieb.kobe-u.ac.jp}

\begin{abstract}
    The Anna Karenina Principle (AKP) holds that success requires satisfying a small set of essential conditions, whereas failure takes diverse forms. We test AKP, its reverse, and two further patterns described as ordered and noisy across novels, online encyclopedias, research papers, and movies. Texts are represented as sequences of functional blocks, and convergence is assessed in transition order and position. Results show that structural principles vary by medium: novels follow reverse AKP in order, Wikipedia combines AKP with ordered patterns, academic papers display reverse AKP in order but remain noisy in position, and movies diverge by genre. Success therefore depends on structural constraints that are specific to each medium, while failure assumes different shapes across domains.
\end{abstract}
\begin{document}

\flushbottom
\maketitle

\section{Introduction}

    This study aims to quantitatively determine, at corpus scale, which structural principles best characterize written communication. We analyze four domains—novels, online encyclopedias, research papers, and movies—to test four competing hypotheses: the Anna Karenina Principle (AKP), which holds that success arises from the simultaneous satisfaction of a small set of essential conditions while failure branches into diverse modes; the reverse AKP, which argues that successful forms are diverse whereas failure collapses into a limited set of common patterns; an ordered type, where both high- and low-evaluation texts converge but follow different structural variants; and a noisy type, where neither side converges and evaluation does not align with structure. Although AKP has been discussed across multiple fields, including ecology, economics, and studies of technological innovation~\cite{rushton1999guns,bornmann2011anna}, systematic tests that connect textual organization with evaluative outcomes remain limited. Existing large-scale efforts address specific signals or media, such as emotional-arc regularities and their association with downloads in fiction~\cite{reagan2016emotional} or the use of screenplay text to predict box-office performance~\cite{eliashberg2014assessing}, but they do not provide a general, cross-domain account of how structural variation relates to outcomes.
    
    In discourse processing, researchers report fragmentation in how they represent and assess structure. They employ competing frameworks, pursue heterogeneous parser objectives, and adopt evaluation procedures that do not align across representations~\cite{webber2012discourse,morey2018dependency}. Developers of readability and cohesion toolkits provide extensive batteries of indices~\cite{mcnamara2014automated}, and researchers proposing discourse-aware metrics such as DiscoScore target coherence in specific generation tasks~\cite{zhao2022discoscore}. Yet no one has established a widely adopted metric that captures the structural dimension of how a text is told, applies across genres, and allows direct comparison with outcomes such as reader engagement, citations, or commercial indicators. To address this gap, we introduce a language-agnostic representation of structural sequences and design two complementary convergence metrics for order and position, which allow us to evaluate how structural similarity aligns with external indicators across domains.
    
    Three factors explain why systematic tests remain scarce. First, researchers often rely on frameworks such as section headings, rhetorical‐relation labels, and RST~\cite{webber2012discourse,morey2018dependency,li2022survey}. These frameworks serve useful purposes but depend heavily on language and genre, and they frequently conflate function (what a unit does) with structure (where it is placed), which limits their portability as comparative metrics. Second, macro-structural noise, including document length and repeated motifs, makes it difficult to isolate the structural effects that matter most. Indices of length and cohesion correlate with many other factors, and researchers still face challenges when they evaluate discourse at scale~\cite{mcnamara2014automated,zhao2022discoscore}. Third, many prior studies analyze limited datasets or focus on a single medium, which restricts data diversity and prevents researchers from identifying whether convergence and divergence take the form of ordered patterns, AKP, reverse AKP, or noisy structures~\cite{reagan2016emotional,eliashberg2014assessing}.

    To address this problem, we introduce functional blocks as minimal structural units. Sentences are mapped to cross-linguistic surface cues such as function-word profiles, syntactic dependencies, stop-word rhythm, and affective tone, and unsupervised clustering groups them into functionally homogeneous blocks. The resulting sequence of blocks represents a document’s structural trajectory. We evaluate these trajectories from two complementary perspectives: order, which captures the sequence of block transitions, and position, which captures when transitions occur along the normalized timeline. This design yields language-agnostic measures that directly capture structural constraints without relying on content words.
    
    We apply this framework to four domains—novels, online encyclopedias, research papers, and movies—using domain-specific evaluation signals such as bookmarks, page views, citations, and box-office revenue. The analysis shows that no single principle governs structural convergence across domains. Novels display reverse AKP in order and weak ordering in position. Wikipedia combines AKP with ordered patterns. Research papers in arXiv align with reverse AKP in order but remain noisy in position due to the IMRAD template. Movies scatter in order overall but diverge by genre, with mystery films showing AKP, science fiction showing reverse AKP, and others remaining noisy; in position, however, successful films concentrate decisive transitions later in the narrative.
    
    Taken together, these findings indicate that structural alignment with evaluation depends on both dimension and medium. Information-oriented texts such as Wikipedia and research papers hinge on order, whereas narrative-oriented texts such as novels and movies hinge on position. External templates like IMRAD can suppress variation and eliminate structural differentiation. Rather than supporting a single universal principle, the results show that success depends on medium-specific structural constraints, while failure takes distinct forms across domains.

    \section{Methodology}
    
    This section presents a three-stage pipeline that maps sentence-level functions to document-level structure and relates those structures to external evaluations across media. We operationalize functions from grammatical and pragmatic cues, build structural representations that separate order from position, and analyze how these representations align with evaluation signals.
    
    \subsection{Measurement framework}
    As shown in Fig.~\ref{fig:overallanalysis}a, we formalize a three-stage framework linking (i) sentence-level functions, (ii) document-level structure, and (iii) external evaluation. For illustration only, Fig.~\ref{fig:overallanalysis}a uses City Lights (1931) and Freytag’s five stages; these are not used in analysis. Structural units in this framework are multi-sentence aggregates that arise from recurring functional signals.
    
    We define function as a sentence-level role inferred from grammatical and pragmatic surface cues, rather than the character- or event-based roles in classical narrative theories (e.g., Propp; Barthes). To obtain language-portable signals and avoid conflating function with structure, we estimate three broad functional types—expository, actional, and emotive—from features such as function-word rates, syntactic-dependency patterns, discourse markers, and affect indicators. This design addresses fragmentation in discourse representations and evaluation procedures and complements readability/cohesion indices that target different constructs~\cite{webber2012discourse,morey2018dependency,mcnamara2014automated,zhao2022discoscore}.
    
    We aggregate consecutive sentences into functionally homogeneous blocks via unsupervised clustering and represent each document as a sequence of block labels. To separate order and position from document length and repeated motifs, we apply run-length compression to the label sequence. We then quantify two complementary facets of structural similarity: (i) order convergence, using edit distance with medoid clustering on compressed sequences; and (ii) position convergence, by recording all block-transition locations, normalizing them by document length, fitting kernel-density estimates, and computing Wasserstein distances between groups. These measurements target structural arrangement rather than content words.
    
    For evaluation mapping, we assign documents to evaluation groups using medium-appropriate external indicators (e.g., box-office revenue for film), treating these as observable proxies for collective reception. Documents are then partitioned into ten evaluation groups based on normalized scores for downstream comparisons. This enables corpus-scale tests relating structural similarity to outcomes without subjective annotation. This enables corpus-scale tests relating structural similarity to outcomes without subjective annotation.
    
    \subsection{Preparation 1 - Domain-specific segmentation and feature extraction}
    We segment each narrative into predefined textual units according to domain conventions: chapters for novels, subheadings for Wikipedia, subsubsections for academic papers, and timestamp-based blocks for subtitles. These units are chosen to reflect local coherence and to capture shifts in rhetorical or narrative function.
    
    From each segment, we extract four feature families designed to characterize structural function while minimizing reliance on semantic content. Part-of-speech (POS) frequency captures grammatical composition and rhythm; stopword frequency indicates discourse-level cohesion and textual transitions; dependency-label frequency encodes syntactic roles that shape logical flow; and emotion scores—computed from polarity and intensity using a sentiment lexicon~\cite{chen2017adversarial}—estimate affective tone. Grammatical and syntactic features are derived using spaCy, and affective features are obtained via lexicon-based aggregation. All features are normalized to probability distributions within each block to ensure comparability across varying lengths. These choices draw on stylometry and discourse analysis~\cite{pennebaker2011secret,boyd2020narrative} and aim to represent the functional form of segments rather than their semantic or topical content.
    
    \subsection{Preparation 2 - Unsupervised functional clustering and sequence representation}
    Within each domain, segment-level feature vectors are clustered with $k$-means to uncover latent structural functions. The resulting clusters represent abstract functional roles that may correspond to phases such as exposition, rising tension, or resolution. The method does not rely on predefined semantic labels, enabling structure-oriented analysis that is unsupervised and data-driven.
    
    We set the number of clusters to $k=5$. This choice is not intended to enforce a single theory of narrative structure; it is inspired by Barthes’ five codes—proairetic, hermeneutic, cultural, semic, and symbolic—as a conceptual scaffold for capturing the distribution and diversity of narrative meaning~\cite{barthes1970sz}. Although finer-grained typologies such as Propp’s 31 functions~\cite{propp1968morphology} are well known, there is no consensus on the optimal granularity for functional segmentation in natural language texts. Five clusters balance computational efficiency, interpretability, and structural coherence.
    
    Clusters are interpreted as functional roles within the narrative. The sequence of functional blocks indicates how the narrative progresses through different functional states and serves as a proxy for structural transitions. Our approach aligns with Rhetorical Structure Theory (RST) in treating semantic and functional structures within a unified layer, but differs by focusing on sequential transitions of functions to enable a linear, temporally oriented analysis. Order alone, however, is not sufficient. In film, for example, whether the shift from exposition to a climactic moment occurs early or late can influence audience response. Accordingly, we analyze structure not only by sequence but also by positional information along the narrative timeline.
    
    \subsection{Preparation 3 - Structural order and positional modeling}
    To interpret coherent sets of functional blocks as narrative structural elements, we approximate higher-level structural blocks by aggregating adjacent functional blocks with overlap allowed. Concretely, we adopt 2-grams of adjacent functional blocks as the minimal unit of structural segmentation. This achieves two aims: (1) it preserves some functional diversity within each structural element, and (2) it makes transitions explicit. Given the absence of a settled framework for the minimum number of functional blocks required to instantiate structural meaning, 2-grams are a pragmatic and computationally efficient choice, consistent with n-gram practice in language modeling~\cite{Jurafsky2000}.
    
\begin{figure*}[!t]  
        \centering
        \includegraphics[width=\linewidth]{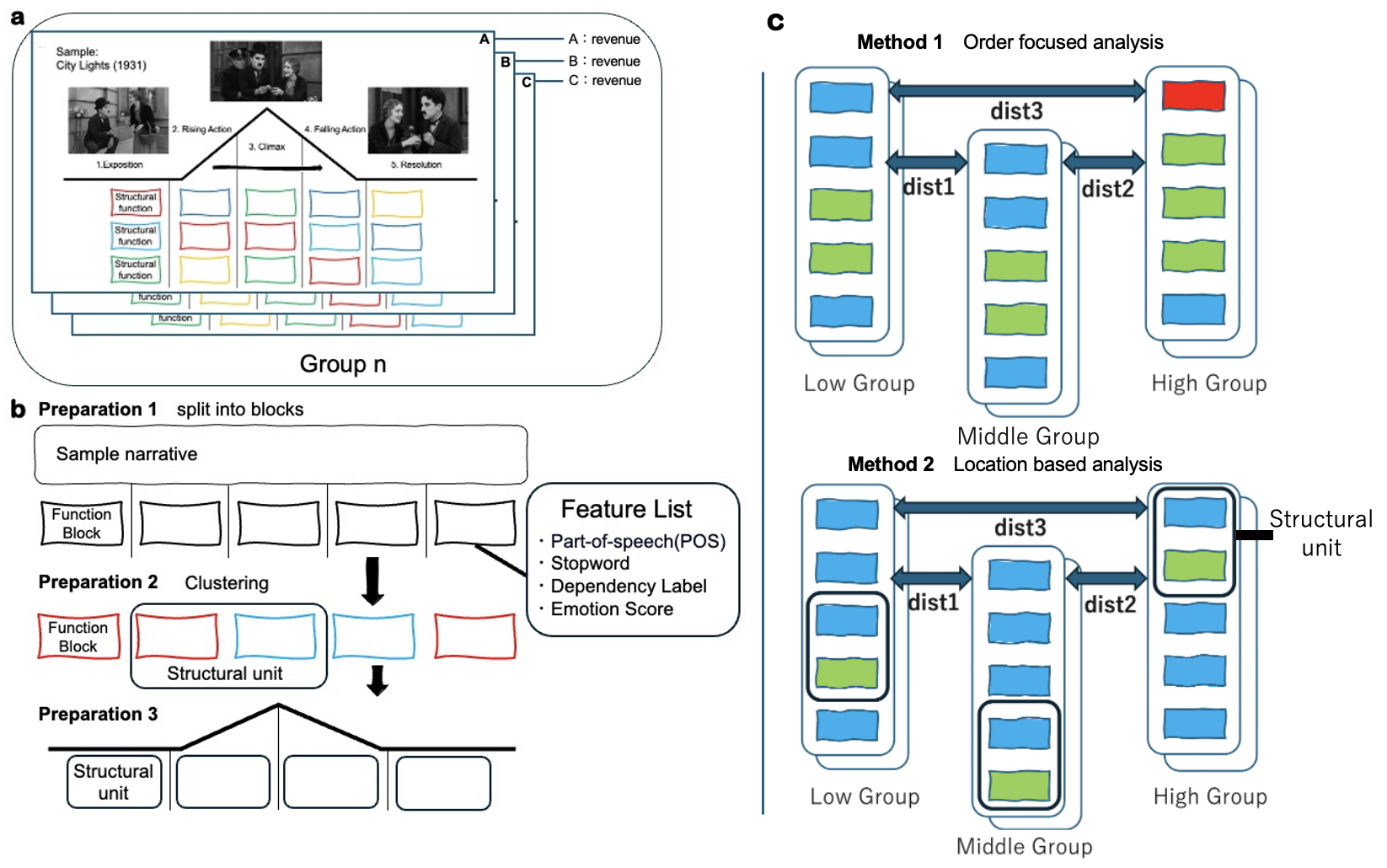}
        \caption{
            Overview of the Proposed Method. (a) An interpretation of Charlie Chaplin’s film City Lights (1931) using Freytag’s Pyramid, a narrative structure model that divides a story into five phases. The accompanying image was generated using ChatGPT-4o. The five phases are as follows: Exposition—the tramp Charlie meets a blind flower girl; Rising Action—the girl mistakenly believes Charlie is a wealthy man, prompting him to raise money for her eye surgery; Climax—Charlie is arrested after being falsely accused of theft, but manages to hand over the money, and the girl's sight is restored; Falling Action—Charlie is released from prison years later; Resolution—the girl, now able to see, realizes Charlie was her true benefactor, culminating in a moving reunion. This study treats narrative structure as a composition of multiple structural functions forming its constituent elements (b) A schematic overview of the proposed method for extracting structural functions and mapping them to narrative components. In Preparation 1, the story is segmented into units. Based on functional features contained in each unit, clustering is performed to associate each cluster with a narrative function (Preparation 2). These clusters are then grouped into functional blocks. To extract narrative components from these functional blocks, n-gram sequences of blocks are constructed and treated as narrative elements. This allows the story to be interpreted as a sequence of higher-order structural elements derived from functional units. (c) The results of two analyses linking the extracted structural functions and narrative elements to evaluation scores. Method 1 focuses on the sequence of structural transitions. Stories are divided into ten groups based on evaluation scores, and differences in the order of structural transitions (clusters) are compared using edit distance. Method 2 focuses on the position of transitions, analyzing where in the story structural shifts occur and comparing their distribution across evaluation groups.}
        \label{fig:overallanalysis}
    \end{figure*}

    We examine the relationship between the derived structural positions and narrative evaluation—such as audience reception—as part of a location-based analysis (Fig.~\ref{fig:overallanalysis}c). Specifically, we compare (i) differences in the order of structural transitions across evaluation groups using edit distance, and (ii) differences in the distribution of where structural shifts occur along the narrative timeline.
    
    \paragraph{Method 1: Transition order}
    As shown in Fig.~\ref{fig:overallanalysis}c (Method 1), we represent each text as an ordered sequence of cluster-assigned functional blocks and apply run-length encoding to collapse consecutive identical clusters into a single transition. Within each evaluation group, we perform $k$-medoids clustering on the compressed sequences to identify representative transition patterns (medoids). We then compute Levenshtein (edit) distances between the medoids of different evaluation groups to quantify how structural order differs across groups. Edit distance treats structural development as a symbolic trajectory and counts the insertions, deletions, and substitutions required to transform one trajectory into another, capturing localized differences and their effect on the overall sequence.
    
    \paragraph{Method 2: Transition position}
    As shown in Fig.~\ref{fig:overallanalysis}c (Method 2), we define a structural unit as a transition between two adjacent functional blocks (a 2-gram). For a sequence such as “AAABBAACCC,” contiguous blocks (“AAA,” “BB,” “AA,” “CCC”) yield cross-boundary transitions “AB,” “BA,” and “AC,” which we treat as structural units. For each text, we record the locations of these transitions, normalize by the total number of segments to obtain values in $(0,1]$, and model the distribution of transition positions within each evaluation group using kernel density estimation. This allows us to detect patterns such as a concentration of transitions toward the end of higher-evaluation works versus more dispersed transitions in lower-rated texts. We quantify differences between groups with the Wasserstein distance.

    \subsection{Data}
    For online novels, we utilized a corpus of 6,580 full-length works sourced from “Shōsetsuka ni Narō,” a major Japanese online fiction platform. Each novel was segmented by chapter, and the number of bookmarks per chapter was used as a proxy for reader evaluation. This dataset reflects a focus on narrative progression and long-term engagement.
    
    Wikipedia articles were drawn from the 2023 English Wikipedia dump, comprising a total of 3,270,926 entries. Texts were segmented based on subsection headings, and their evaluation metric was defined by the annual page view count. As an encyclopedic medium, Wikipedia provides insight into how structural elements contribute to reader attention in non-narrative contexts.
    
    The academic paper corpus consisted of 720,126 papers from arXiv, covering a wide range of disciplines including physics, mathematics, and computer science. These texts were segmented at the subsubsection level, and citation counts normalized over a ten-year window were used as the evaluation metric. In this domain, the structure primarily serves to support logical progression and scholarly communication.
    
    Lastly, we examined 199,754 movie subtitle files obtained from OpenSubtitles.org, which were linked to box office performance and genre metadata from the Full TMDB Movies Dataset 2024. Segmentation was conducted using the Bayesian Blocks algorithm \cite{scargle2013bayesian}, which identifies optimal breakpoints based on changes in timestamp intervals. Subtitle download counts were adopted as the evaluation measure, reflecting user interest and accessibility.
    
    To ensure consistency in structural granularity, we filtered out texts with an exceptionally high or low number of segments using the interquartile range (IQR). The remaining texts were then grouped into ten bins (group\_0 to group\_9) based on rank-normalized evaluation scores.

    \begin{figure*}[!t]  %← figure* では h が使えないので !t か !p
        \centering
        \includegraphics[width=\linewidth]{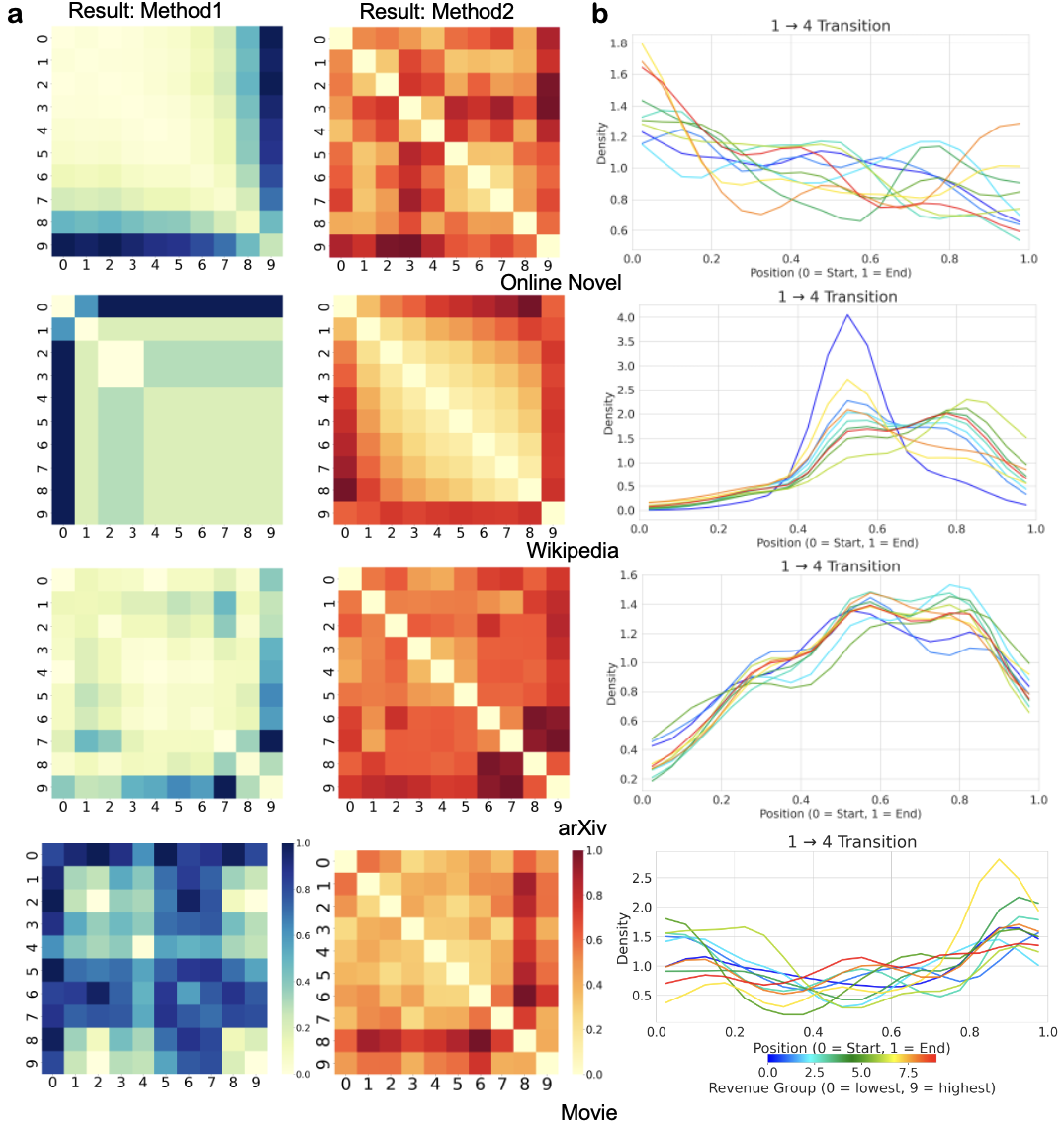}
        \caption{(a) Heatmap showing pairwise structural differences among evaluation groups, computed using the methods outlined in Method 1 (transition order) and Method 2 (transition position). Higher group numbers (with Group 9 being the highest) correspond to higher evaluation scores. Color intensity the magnitude of structural dissimilarity, with darker shades indicating greater divergence in structural patterns between groups. (b) Positional distribution of the 1 → 4 structural transition across evaluation groups, obtained with the extended procedure described in Method 2. The horizontal axis shows the normalized position in the text (0 = beginning, 1 = end), and the colour gradient encodes evaluation level from cool (low-evaluation groups) to warm (high-evaluation groups). The plot allows direct comparison of where this representative transition tends to occur in texts of differing quality. Because cluster indices are assigned independently for each dataset, the label “1 → 4” does not imply an identical functional change across media.
            }
        \label{fig:result1}
    \end{figure*}

    \begin{figure*}[ht]
        \centering
        \includegraphics[width=\linewidth]{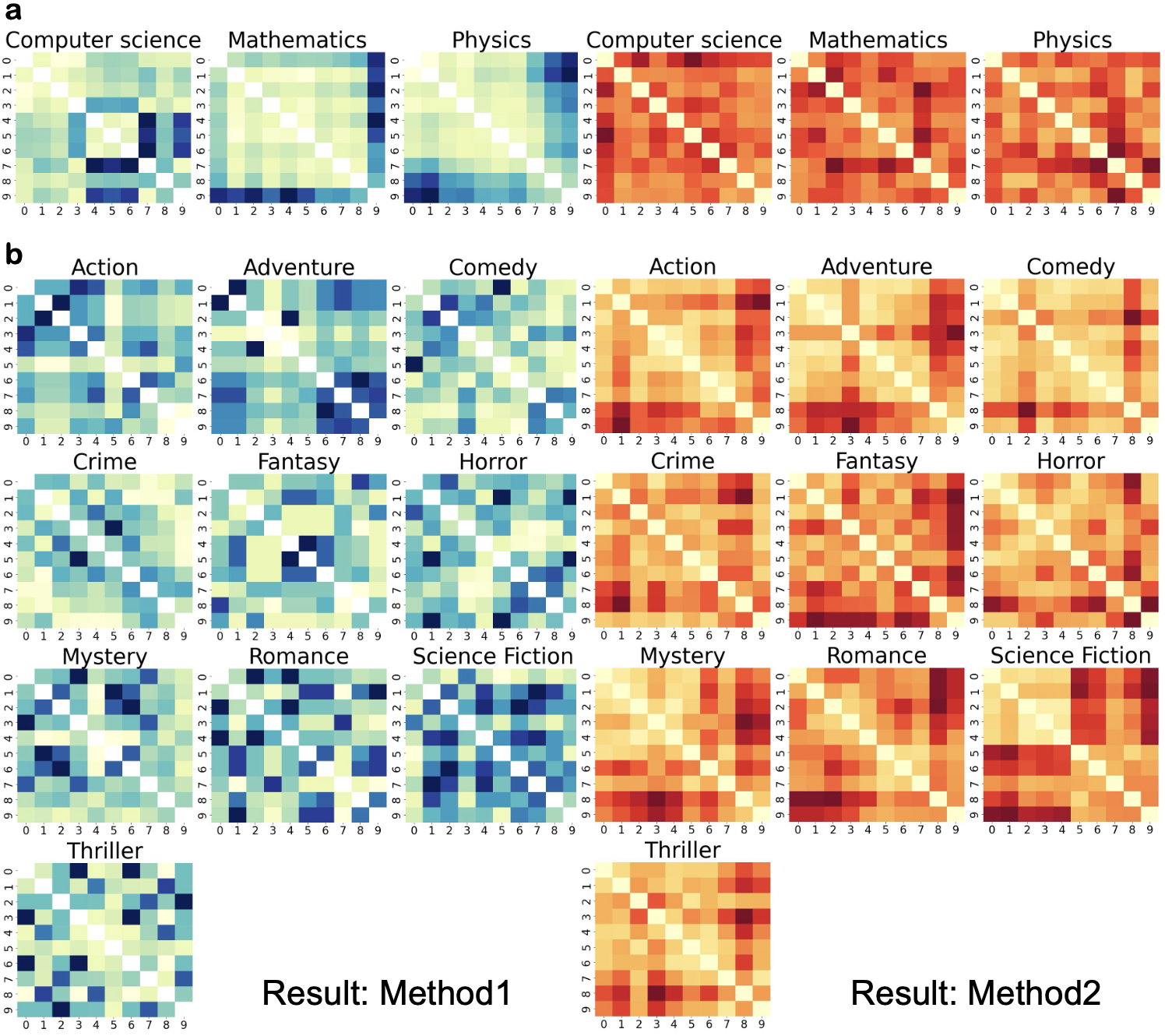}
        \caption{Genre-wise structural analysis for arXiv and movie datasets. a presents the results of Method 1 and Method 2 applied to the arXiv dataset, disaggregated by research field. Only categories with more than 10,000 samples were included in the analysis. b shows the corresponding results for movie genres, again limited to genres with over 10,000 samples. In the case of movies, individual works often have multiple genre tags; such works were counted in all relevant genre categories. This approach allows for the identification of cross-genre structural tendencies and the visualization of patterns in multi-genre content.
            }
        \label{fig:result2}
    \end{figure*}

    \section{Results}
    
    The results of this study can be interpreted through four structural hypotheses: ordered, AKP, reverse AKP, and noisy. Figures~\ref{fig:result1} and \ref{fig:result2} present the distance matrices across media and by discipline/genre, illustrating how each hypothesis manifests. Each hypothesis reflects a distinct alignment or misalignment between evaluation outcomes and structural similarity. In our interpretation, we rely on the heatmaps of transition order (Method 1) and transition position (Method 2). Order reflects the similarity of sequential clusterings, while position reflects the similarity of normalized placements along the timeline. The 1$\rightarrow$4 transition distributions in Figure~\ref{fig:result1}b further serve as supplementary evidence for positional differentiation.
    
    \subsection{Ordered type}
    The ordered type arises when both low- and high-evaluation groups converge internally, but follow different structural variants. Evaluation outcomes depend on which variant is followed: one structural path consistently aligns with low evaluation, while another aligns with high evaluation.  
    
    This pattern was especially visible in the transition position heatmaps. In Figure~\ref{fig:result1}a, Wikipedia showed a clear ordered structure: high-evaluation groups (7 to 9) formed a tight light-colored block, while low groups (0 to 2) clustered separately, with darker bands marking cross-group distances. Mid-range groups (3 to 6) occupied an intermediate zone, indicating gradual structural shifts along the evaluation axis. Online novels also exhibited an ordered pattern in position, though less strongly: high-evaluation groups converged loosely, with weaker separation from low groups. Movies displayed a similar ordered structure in position: high-grossing films concentrated transitions toward the latter part of the normalized timeline, while low-grossing films aligned earlier, producing a distinct split between high and low groups despite genre heterogeneity. By contrast, arXiv showed no ordered separation in position: the heatmap remained uniformly light, reflecting the rigid IMRAD format that enforces fixed rhetorical placements.  
    
    The 1$\rightarrow$4 transition distributions in Figure~\ref{fig:result1}b reinforced these observations: in Wikipedia and movies, high-evaluation groups shifted peaks toward the end of the timeline, while low groups clustered earlier. Novels displayed the same trend but with broader peaks, consistent with a weaker ordered pattern. In arXiv, distributions overlapped almost entirely, providing no evaluative differentiation.
    
    \subsection{AKP}
    The AKP type occurs when high-evaluation groups converge on a narrow structural pathway, while low-evaluation groups remain scattered. This corresponds to the Anna Karenina Principle: success requires fulfilling specific conditions, whereas failure can take many forms.  
    
    Wikipedia provided the clearest AKP example in order (Figure~\ref{fig:result1}a). High-evaluation groups (7 to 9) formed a coherent light block, while low groups, particularly Group 0, remained dispersed, with darker cross-group distances. Position heatmaps also revealed AKP-like convergence: high groups clustered tightly, separated from scattered low groups, making Wikipedia simultaneously ordered and AKP in nature.  
    
    At the genre level, mystery films displayed AKP in order (Figure~\ref{fig:result2}b). High-grossing mysteries formed a coherent trajectory, while low-grossing ones remained dispersed. The 1$\rightarrow$4 transition distributions (Figure~\ref{fig:result1}b) further supported this: Wikipedia and movies showed high-evaluation groups concentrated in later transitions, underscoring the narrow structural pathways associated with success.
    
    \subsection{Reverse AKP}
    The reverse AKP type arises when low-evaluation groups converge tightly on repetitive templates, while high-evaluation groups diverge across multiple trajectories. This inverts the AKP intuition: failure is homogeneous, while success is achieved through diverse paths.  
    
    Online novels clearly exemplified reverse AKP in order (Figure~\ref{fig:result1}a). Low groups (0 to 2) formed a tight light block, reflecting conventional sequential patterns, while high groups (7 to 9) spread broadly. arXiv also showed reverse AKP in order: field-level matrices (Figure~\ref{fig:result2}a) revealed that low-evaluation papers, particularly in mathematics and physics, converged on formulaic orderings, while high-evaluation papers diverged across multiple alternatives. Among movies, science fiction (Figure~\ref{fig:result2}b) showed the same pattern: low-revenue films followed repetitive templates, while high-revenue films branched into diverse trajectories.  
    
    Thus, reverse AKP highlights structural asymmetry: failures are repetitive and homogeneous, whereas successes allow for multiple viable trajectories.
    
    \subsection{Noisy type}
    The noisy type occurs when neither low- nor high-evaluation groups converge, with both remaining dispersed. Here, structural similarity does not align with evaluation outcomes, either because conditions for success are absent or other modalities dominate.  
    
    arXiv in position provided the strongest example (Figure~\ref{fig:result1}a): the heatmaps were uniformly light, with little difference across groups. The IMRAD format suppresses variation in rhetorical positioning, erasing evaluative differentiation. Movies also fell into the noisy type in order: both high- and low-grossing films scattered without forming stable clusters (Figure~\ref{fig:result1}a). Genre-level results (Figure~\ref{fig:result2}b) reinforced this for comedy and romance, which lacked evaluative convergence.  
    
    The 1$\rightarrow$4 transition distributions (Figure~\ref{fig:result1}b) confirmed these findings: in arXiv, curves overlapped across groups, eliminating differences; in movies, order-based distributions also failed to distinguish evaluation.
    
    \subsection{Summary}
    Across media, no single principle explained alignment between evaluation and structure. Instead, patterns varied by dimension (order vs.\ position) and medium. Online novels showed reverse AKP in order and weak ordering in position. Wikipedia exhibited AKP in order and ordering in position. arXiv aligned with reverse AKP in order and noisy patterns in position. Movies were noisy in order overall, but genres diverged: mystery aligned with AKP, science fiction with reverse AKP, and others with noisy patterns; in position, films tended toward ordering, with high-evaluation groups clustering later in the timeline.  
    
    Taken together, Figures~\ref{fig:result1} and \ref{fig:result2} demonstrate that evaluation depends on how structure operates in each medium: order plays a central role in information-oriented texts (Wikipedia, arXiv), while position dominates in narrative-oriented texts (novels, movies), and strong external templates like IMRAD can suppress evaluative differentiation altogether.
    
    \section{Discussion}
    
    This study investigated how structural coherence, defined by the order and the position of transitions between functional text segments, relates to external evaluation across four textual media: online novels, Wikipedia articles, academic papers, and movie subtitles. We employed two complementary methods to uncover these structural patterns. Method~1 focuses on the sequence of transitions between functional clusters, referred to as transition order. Method~2 examines the timing of those transitions within a length-normalized representation of each text, referred to as transition position. Our operationalization aligns with discourse theories in which coherence arises from relations among spans as in Rhetorical Structure Theory and from stable referential progression as in entity-based models, which jointly motivate attention to both order and timing \cite{mann1988rhetorical,barzilay2008modeling}.
    
    Across all datasets, structural characteristics were meaningfully associated with evaluation outcomes, yet the nature and strength of the association varied by medium. In online novels, both transition order and transition position correlated with reader evaluation. High-evaluation novels consistently followed distinct and cohesive transition sequences and they placed key structural shifts at deliberate moments. This observation accords with large-scale narrative analyses showing that successful stories concentrate on a limited set of recurrent emotional arcs, some of which align with stronger audience response \cite{reagan2016emotional}. The highest-evaluation group, Group~9, diverged clearly from mid- and low-evaluation groups, which indicates that well-organized development is a hallmark of successful narratives in this domain.
    
    In Wikipedia, structural patterns showed a different trend. Method~1 revealed that the lowest evaluation group, Group~0, diverged substantially in transition order from all other groups, suggesting weaker adherence to expected rhetorical order. This matches studies showing that structural and size-related features such as sectioning, references, and word count are strong predictors of page quality \cite{blumenstock2008size}. Method~2 further showed that the timing of structural transitions correlates with evaluation: high-evaluation articles concentrate transitions near the middle of the text, whereas low-evaluation articles display more dispersed placement. Because article length correlates with assessed quality, part of this positional effect may be mediated by length, a known confound in quality modeling \cite{blumenstock2008size}.
    
    The arXiv corpus showed that domain norms shape the link between structure and evaluation. Mathematics and Physics exhibited systematic differences in transition order between high- and low-citation groups, while Computer Science, which follows the IMRAD convention more rigidly, showed limited variation. These findings cohere with historical and descriptive accounts of the widespread adoption of IMRAD and with rhetorical zoning studies that constrain where conceptual moves typically occur \cite{sollaci2004introduction,teufel2002summarizing}. Regardless of field, transition position remained largely stable across evaluation groups, which indicates that templates impose strong constraints on where shifts occur in academic discourse.
    
    In movie subtitles, transition order showed little consistent association with evaluation. Method~2 nevertheless revealed that the timing of structural shifts linked to climactic content aligned with viewer reception. Top-grossing films concentrate key transitions toward the latter part of the narrative, in line with cognitive-narrative accounts that document late-climax and multi-part structures in popular cinema \cite{cutting2016narrative}. The fact that this timing effect emerges from dialogue alone, despite the multimodality of film, suggests that the macro-structural placement of major events leaves a detectable signature in text.
    
    Taken together, our results furnish corpus-scale support for the Anna Karenina Principle in written communication. Across the four media, successful texts judged by bookmarks, page views, citations, or box-office revenue converge on a narrow corridor of structural coherence, whereas unsuccessful texts deviate from this corridor in many ways \cite{bornmann2011anna}. Which structural dimension is essential depends on the medium. Narrative-driven media such as online fiction and film are evaluated primarily through the timing of functional transitions: decisive shifts placed late, after adequate build-up, consistently coincide with higher reception. Information-oriented genres, by contrast, reward adherence to a stable rhetorical order. Wikipedia articles and mathematics or physics papers with the highest evaluations follow predictable sequences of moves, whereas low evaluations correlate with disordered or misplaced transitions. Satisfying or violating a single temporal or sequential constraint can tip a text from “happy” to “unhappy,” consistent with Tolstoy’s asymmetry.
    
    This medium dependence also invites a temporal reinterpretation of classical structural models. Frameworks such as Freytag's Pyramid and IMRAD enumerate the parts of a text, yet our results indicate that their impact depends less on the elements themselves than on when those elements surface relative to the reader's cognitive timeline. This view is compatible with research on situation models and event indexing in comprehension, which shows that readers track changes along dimensions such as time, causation, space, goals, and protagonists; structural shifts that align with these expectations are easier to integrate \cite{zwaan1998situation}. In this sense, structure functions as a distribution of functional shifts aligned with genre-specific pacing. Texts succeed when they introduce changes at points that are sufficiently predictable to be coherent yet sufficiently varied to remain engaging.
    
    Limitations and future work. Our approach identifies functional blocks by unsupervised clustering of surface-level features such as part-of-speech profiles, dependency patterns, and sentiment, and it models coherence through adjacent bigram transitions in order and position. This design promotes cross-domain applicability but may miss higher-level discourse organization, including long-range dependencies, thematic revisit, and hierarchical rhetorical relations such as Cause, Contrast, and Elaboration \cite{barzilay2008modeling,teufel2002summarizing}. Future research should integrate coreference chains, discourse markers, and annotated rhetorical relations, and represent texts with multi-layer graphs or hierarchical models to capture multi-scale structure. A second limitation concerns the external signals for evaluation. Bookmarks, page views, citations, and box-office revenue are proxies shaped by visibility, recency, and platform algorithms. Although normalization and binning mitigate some of these factors, causal interpretation remains limited. Controlled human studies that manipulate order or position, together with multimodal analyses for film that align text with audio and visual cues, will help validate the mechanisms and test generalizability across languages and cultures.

\end{document}